A Generic Stochastic Hybrid Car-following Model Based on Approximate Bayesian Computation

## Author Information

Jiwan Jiang[1,3], Yang Zhou[2], Xin Wang[3], Soyoung Ahn[1]*

Affiliations

[1]**Department of Civil and Environmental Engineering, University of Wisconsin-Madison, Madison, WI, USA**

[2]**Zachry Department of Civil and Environmental Engineering, Texas A&M University, College Station, TX, USA**

[3]**Department of Industrial and Systems Engineering, University of Wisconsin-Madison, Madison, WI, USA**

Corresponding author

Correspondence to: Soyoung (Sue) Ahn
ORCID:0000-0001-8038-4806

# Abstract

Car following (CF) models are fundamental to describing traffic dynamics. However, the CF behavior of human drivers is highly stochastic and nonlinear. As a result, identifying the "best" CF model has been challenging and controversial despite decades of research. Introduction of automated vehicles has further complicated this matter as their CF controllers remain proprietary, though their behavior appears different than human drivers. This paper develops a stochastic learning approach to integrate multiple CF models, rather than relying on a single model. The framework is based on approximate Bayesian computation that probabilistically concatenates a pool of CF models based on their relative likelihood of describing observed behavior. The approach, while data-driven, retains physical tractability and interpretability. Evaluation results using two datasets show that the proposed approach can better reproduce vehicle trajectories for both human-driven and automated vehicles than any single CF model considered.

# 1. Introduction

Car-following (CF) behavior describes how one vehicle follows its nearest preceding vehicle. This fundamental driving behavior is deeply linked to system-level behavior such as traffic dynamics (i.e., spatial-temporal evolution of traffic) and has important implications for traffic safety, energy consumption, and emissions. A CF model for human driven vehicles (HDV) was first introduced by Pipe(Pipes, 1953) in the 1950s. Today a wealth of CF models exists in the literature, including stimulus-response type models (Gazis et al., 1959; Herman et al., 1959), Newell's simplified CF model (Newell, 1961) and its extensions (Chen et al., 2014; Laval & Leclercq, 2010), statistical physics-based models such as optimal velocity model (OVM) (Bando et al., 1995) and intelligent driver's model (IDM) (Kesting et al., 2010), and psycho physics based models such as Wiedemann model (Wiedemann, 1974). Notably, some of these models have been incorporated in various commercial microsimulations. For a detailed review of CF models, see Saifuzzaman & Zheng, 2014).

A plethora of CF models gave rise to persistent debates about which model best describes the real-world CF behavior. These debates continue today despite the nearly 70 years of history. These debates stem from the complexity of CF behavior, which is intrinsically nonlinear, heterogeneous, and stochastic. It has been challenging to replicate observed data with a single CF model, given that most existing CF models have deterministic formulations. Some exceptions exist to address the complexity in CF behavior through (1) probabilistic distributions of CF model parameters (Higgs & Abbas, 2015; Kerner, 2004; Treiber et al., 2010) and (2) multi-regime CF models according to traffic conditions (Kerner, 2004; Kidando et al., 2020; Treiber et al., 2010). The former approach, however, is typically parametric, requiring an assumption of a theoretical distribution. This can lead to bias when there is a discrepancy between the assumed and true distributions. For the latter approach, different CF models are considered for different traffic conditions. These frameworks, however, are deterministic and consider a relatively narrow selection of CF models (e.g., 3-4 models). Some data-driven methods such as clustering (Higgs & Abbas, 2014) and regression (Papathanasopoulou & Antoniou, 2015) are considered to characterize stochasticity; however, they provide little physical interpretation.

The emergence of automated vehicles (AVs) brings another level of complexity to traffic flow systems. In academic literature, AV CF control algorithms have been predominantly developed based on the principles of control theory (e.g., linear feedback (Makridis et al., 2021; Zhou et al., 2019), model predictive control (MPC) (Shi & Li, 2021 , Zhou et al., 2020), or artificial intelligence (Shi et al., 2021), distinct from the mathematical and physical approaches of the CF models of HDVs. Thus, the AV CF behavior could be different from the HDV behavior. Further, similar to HDVs, AV CF can be affected by actuation delay, uncertain vehicle dynamics, road conditions, and traffic conditions, leading to highly stochastic behavior. Finally, AVs manufactured by different car companies are available on the market today. Their control algorithms are likely different, yet unknown to the public, which hinders our ability to characterize the CF behavior of AVs.

To better understand the CF behavior of AVs, several field experiments involving vehicles with adaptive cruise control (ACC) have been conducted (Li et al., 2022; Makridis et al., 2021; Shi & Li, 2021). The data from these experiments have been used by several studies to model and replicate the AV CF behavior, with two different approaches: (1) model-based and (2) data-driven. In the model-based approach, a CF model is assumed, and its parameters (and their distributions) are learned from observations. This approach readily offers physical interpretations of the behavior but suffers from potential model mismatch. Further, efforts to capture stochasticity in CF behavior through estimating parameter distributions typically involve an assumption of distribution (Rahman et al., 2015). Thus, a mismatch in CF model and/or the parameter distributions can compromise the learning results and the descriptive power of the assumed CF model. In addition, learning the stochastic behavior with non-analytical CF model (e.g., MPC) is computationally demanding, and thus an efficient tool is necessary. In contrast, data-driven methods, such as neural network (NN) based methods (Hornik et al., 1989, 1990), are capable of describing any type of nonlinear functions given sufficient neurons and layers. However, the black-box nature of these methods hinders direct physical interpretation. Further, the learned NN is limited by the training dataset, and thus, it may not effectively handle corner cases not represented in the training data.

The review above reveals the persistent challenges to address highly nonlinear and stochastic nature of CF behavior that has been further complicated by the arrival of AVs. To fill this major gap, this paper presents a comprehensive framework that systematically considers a pool of CF models and various uncertainties and stochasticity. Specifically, the proposed framework generates a hybrid CF model that represents the probabilistic concatenation of a pool of CF models based on their abilities to reproduce the real behavior measured from sensors. The general framework is illustrated in Fig. 1. The core method of the framework is approximate Bayesian computation (ABC), a computational method to approximate the posterior model parameter distributions through simulations without assuming a specific likelihood function (Toni et al., 2009). ABC has been originally used in population genetics (Beaumont et al., 2002; Tavaré et al., 1997), but has also been widely applied in biology (Liepe et al., 2014) and ecology (DiNapoli et al., 2021). Our recent study, Zhou et al. (2022), developed a methodology based on ABC to calibrate a single CF model or controller in a stochastic fashion. This approach serves as a foundation for the present work that probabilistically compares across different CF models and generates a stochastic hybrid model.

In our framework, particles (i.e., sets of model parameter values) for each CF model are randomly generated in large quantity from an assumed prior joint distribution in an independent fashion. When only a single CF model is considered, all accepted particles from the model can be used to construct the posterior distribution of the model parameters. In contrast, when multiple CF models are considered, particles are evaluated based on the universal distance function (across CF models) that measures the discrepancy between simulated vehicle trajectories based on the particles and real trajectories. A universal threshold for the distance is then applied to accept only the particles that generate trajectories within the acceptable distance. Accordingly, the relative share of accepted particles represents the relative likelihood of the model describing the observed behavior. Then the accepted particles are used to approximate the posterior distribution of the hybrid CF model in a Bayesian fashion by concatenating the models according to the relative likelihoods. Thus, the learned hybrid CF model enhances the capability of describing nonlinear CF behavior while preserving the physical meaning of each CF model.

Note that the proposed framework is stochastic and hybrid, designed to provide a richer understanding of CF behavior while improving learning accuracy. It is *stochastic* in the senses that (1) it estimates the joint distributions of CF model parameters; and (2) it considers the relative likelihood of each CF model fitting the observed behavior. It is *hybrid* in the senses that (1) it concatenates various CF models, rather than relying on a single best-fitting model, as previously done, according to the relative likelihood; and (2) it deploys a data-driven method to estimate the joint distributions of physics-based CF models, thereby retaining physical interpretability while improving learning accuracy. The hybrid model is particularly useful in determining which control algorithm is most likely adopted for an AV and approximating its behavior in the absence of the controller knowledge. Our method is verified through a series of evaluations using synthetic and real data. The hybrid model is shown to significantly outperform any single model or deterministic models in reproducing vehicle trajectories.

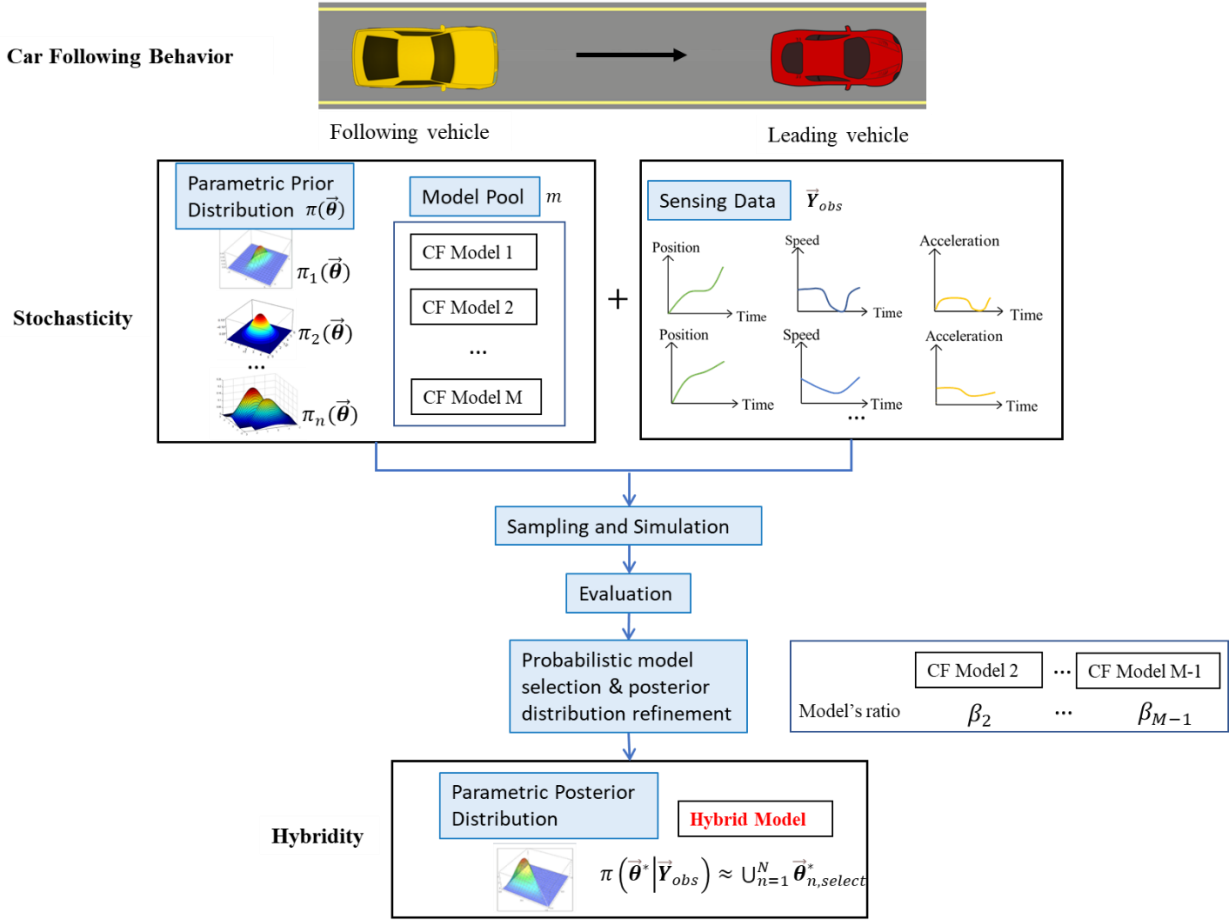

**Fig. 1 Scheme of the CF behavior learning framework.**

## 2. Methodology

### 2.1. General stochastic CF learning problem

A general form of CF learning problem can be described as below:

$$\min_{\vec{\theta}} g(\mathbb{y}_f - \hat{\mathbb{y}}_f) \tag{1a}$$

s.t.
$$\hat{\mathbb{y}}_f = f(\mathbb{y}_l; \vec{\theta}), \tag{1b}$$

where $\mathbb{y}_f$ is a set of observed ground-truth state portfolios of the following vehicles, $\mathbb{y}_f = \{\vec{Y}_{f,1,obs}, \vec{Y}_{f,2,obs} \ldots, \vec{Y}_{f,I,obs}\}$; we define the state portfolio for vehicle $i = 1,2,\ldots,I$, denoted by $\vec{Y}_{f,i,obs} = [\vec{p}_{f,i,obs}, \vec{v}_{f,i,obs}, \vec{a}_{f,i,obs}]^T$, to represent a vector of observed position, speed, and acceleration profiles over time. Similarly, we denote the set of simulated state portfolios for the following vehicles by $\hat{\mathbb{y}}_f = \{\vec{Y}_{f,i,sim}\}_{i=1,2,\ldots,I}$. Here, we extend the operation "−" for the state portfolios set to define $g(\cdot)$ as a predefined error (distance) function, measuring the deviation between the observed and learned state portfolios. Here *Eq.* (1a) is the objective function to measure the goodness of fit. In particular, the CF model, denoted by $f(\mathbb{y}_l; \vec{\theta})$, is parameterized on vector $\vec{\theta}$, given the leading vehicle's state portfolios $\mathbb{y}_l$ $= \{\vec{Y}_{l,i,obs}\}_{i=1,2,...,I}$. According to *Eq.* (1a) and *Eq.* (1b), the error function, CF model, and observed state portfolios are three critical components for model learning.

For stochastic extension, we introduce uncertainty to parameter $\vec{\theta}$. In particular, we revise the CF model into $f(y_l; \pi(\vec{\theta}))$, where $\pi(\vec{\theta})$ indicates a sampled $\vec{\theta}$ from a given random distribution $\pi$. Therefore, instead of finding the best $\vec{\theta}$ value in deterministic CF models, the decision variable for stochastic CF model learning is the whole distribution $\pi(\vec{\theta})$. In addition, to explicitly reflect the variation of CF model forms, e.g., IDM or MPC, we denote $M$ to be the index set of all CF models, and $f(y_l; \pi(\vec{\theta}), m)$ to be one specific CF model $m \in M$. In our context, we further consider a hybrid stochastic CF model learning problem. In particular, stochasticity comes from not only the parameter $\vec{\theta}$ but also the form of the CF model. We denote such a hybrid stochastic CF model by $f\left(y_l; \pi(\vec{\theta}, m)\right)$ to show the model and parameters can be random at the same time.

Unlike the previous studies that rely on a single model, the hybrid stochastic CF model makes use of multiple CF models, providing rich model function representability. Further, the ABC-based model selection framework offers flexibility and interpretability.

*2.2. Approximate Bayesian computation mechanism*

We adopt the ABC-based learning of CF model parameters presented in our previous paper (Zhou et al., 2022) as summarized below. The main focus of Bayesian inference is to obtain the posterior distribution when given observations and the prior distribution of parameters, written as:

$$\pi(\vec{\theta}|\vec{Y}_{obs}) \propto l(\vec{\theta}|\vec{Y}_{obs})\pi(\vec{\theta}), \qquad (2)$$

where $\pi(\vec{\theta})$ represents the prior distribution of parameters, $l(\vec{\theta}|\vec{Y}_{obs})$ represents the likelihood of $\vec{\theta}$ given the observed state portfolio data $\vec{Y}_{obs}$, and $\pi(\vec{\theta}|\vec{Y}_{obs})$ is the posterior distribution proportional to the right-hand side. The core of our CF model learning is to obtain the approximation of the posterior distribution $\pi(\vec{\theta}|\vec{Y}_{obs})$, implying we learned a CF model from the observation.

Although the prior distribution of CF model parameters can be given or assumed, the likelihood function is generally unavailable. Hence, ABC can be applied to approximate the likelihood $l(\vec{\theta}|\vec{Y}_{obs})$ by simulations without assuming a specific function form, relying on large-scale computation (Toni et al., 2009). Such a likelihood-function-free structure renders ABC a powerful tool to learn complicated even non-analytical CF models.

A simple but important ABC approach is the ABC rejection sampling (ABC-RS) (Beaumont et al., 2002). It repeats the following simulation process: (1) randomly sample a parameter vector $\vec{\theta}^*$, called a *particle*, from a given prior distribution $\pi(\vec{\theta})$; (2) plug $\vec{\theta}^*$ into the CF model $f(y_l; \vec{\theta}^*)$ to simulate state portfolios $\hat{y}_f^*$; (3) compare the simulated data against the real observation using a pre-defined distance function $g(y_f - \hat{y}_f^*)$ and accept the particle $\vec{\theta}^*$ if the distance is smaller than a certain threshold. Such distance is called the *score of the particle*. A large number of simulations are typically needed to obtain a sufficient number of accepted particles. Finally, the posterior joint distribution is estimated using the $N$ accepted particles, written as:

$$\pi(\vec{\theta}^*|\vec{Y}_{obs}) \approx \bigcup_{n=1}^{N} \vec{\theta}_{n,select}^* \qquad (3)$$

where $\vec{\theta}^*_{n,select}$, $n = 1, 2, \ldots, N$ is an accepted particle. Without loss of generality, we assume the particles selected are sorted in ascending order based on their score, and the index $n$ indicates the order. The central idea of ABC is that the particles that reproduce state portfolios close to the real observation should also have good proximity to the learned posterior distribution.

Regarding the distance function $g$ in *Eq.* (1a), multiple measures are applied to assess the learning accuracy, such as the sum of squared errors (Toni et al., 2009) and Euclidean distance (DiNapoli et al., 2021). Here, we design our own distance function. We first define the deviations (errors) of vehicle position, ($e_{p,\vec{\theta}^*}$), velocity ($e_{v,\vec{\theta}^*}$), and acceleration ($e_{a,\vec{\theta}^*}$):

$$e_{p,\vec{\theta}^*} = \frac{1}{I}\sum_{i=1}^{I}\|\vec{p}_{f,i,sim} - \vec{p}_{f,i,obs}\|, \tag{4a}$$

$$e_{v,\vec{\theta}^*} = \frac{1}{I}\sum_{i=1}^{I}\|\vec{v}_{f,i,sim} - \vec{v}_{f,i,obs}\|, \tag{4b}$$

$$e_{a,\vec{\theta}^*} = \frac{1}{I}\sum_{i=1}^{I}\|\vec{a}_{f,i,sim} - \vec{a}_{f,i,obs}\|, \tag{4c}$$

where $I$ is the total number all CF pairs chosen for learning. Then the distance function or the score of particle $\vec{\theta}^*$, denoted by $g_{\vec{\theta}^*}$, can be defined as the weighted sum of the error:

$$g_{\vec{\theta}^*} = \alpha_1 e_{p,\vec{\theta}^*} + \alpha_2 e_{v,\vec{\theta}^*} + \alpha_3 e_{a,\vec{\theta}^*}, \tag{5}$$

where $\alpha_1$, $\alpha_2$, and $\alpha_3$ are weights for each error term, respectively, and $\alpha_1 + \alpha_2 + \alpha_3 = 1$. Hereafter, we set $\alpha_1 = 0.5$, $\alpha_2 = 0.3$, $\alpha_3 = 0.2$ by default, as position data are generally most reliable in state portfolios.

In practice, when $I$ is a large number and the simulation of $f(y_i; \vec{\theta})$ is time-consuming (e.g., MPC), the evaluation of $g_{\vec{\theta}^*}$ can be slow. To speed up, we can estimate $e_{p,\vec{\theta}^*}, e_{v,\vec{\theta}^*}, e_{a,\vec{\theta}^*}$ and $g_{\vec{\theta}^*}$ by randomly sampling one CF pair $i^*$, e.g., $e^*_{p,\vec{\theta}^*} = \|\vec{p}_{f,i^*,sim} - \vec{p}_{f,i^*,obs}\|$, denoted by $g^*_{\vec{\theta}^*}$. This down sampling process can be considered as the simulation uncertainty. Note that since $g^*_{\vec{\theta}^*}$ is randomly evaluated, pair selection $i^*$ can potentially dominate the impact of particle $\vec{\theta}^*$. To avoid such an over-representing issue, we select particles based on their corresponding CF pairs, i.e., $\pi(\vec{\theta}^*|\vec{Y}_{obs}) \approx \cup_{i=1}^{I}\cup_{n_i=1}^{N_i}\vec{\theta}^*_{n_i,select}$, where $N_i$ is the number of particles evaluated by CF pair $i$. Similarly, without loss of generality, we assume $n_i$ to be the order of sorted particles evaluated from each CF pair $i$, respectively.

Therefore, the learning result of the stochastic CF model using the ABC method is a distribution estimated by the *optimal particle set* $\Theta^{opt} = \cup_{i=1}^{I}\cup_{n_i=1}^{N_i}\vec{\theta}^*_{n_i,select}$. To reproduce the state portfolios using the learned stochastic CF model, one particle is randomly selected from $\Theta^{opt}$ to capture the uncertain nature of driving behavior.

### 2.3. Hybrid CF model

The above ABC framework can be adopted for a specific CF model form. Note that the learned result takes the form of a set of selected particles. Thus, it can be easily extended to incorporate multiple CF models for learning to enhance its representability. A hybrid model retains a subset of particles across models, which can offer a richer understanding of CF behaviors of HDVs and AVs, and more refined micro-simulation. The detailed steps to obtain hybrid model are described below and shown in Fig. 2.

**Step 1: Initialization**

Define a set of candidate CF models, indexed by $m \in M$, where both CF controllers for AVs and CF models for traditional HDVs are included. For each CF model/controller $m$, we denote a sampled particle by $\vec{\theta}_m^*$ under the given prior distribution $\pi_m(\cdot)$. The prior distribution set for overall models is $\Pi = \{\pi_1(\cdot), \pi_2(\cdot), \dots, \pi_M(\cdot)\}$.

**Step 2: Learning of each model through ABC**

We process ABC-RS independently for each model $m$. A large number of (e.g., >1 million) particles are independently sampled for each model. When all learning processes are completed, the optimal selected particle set for each model is obtained as $\Theta_m(N_m) = \cup_{i=1}^{I} \cup_{n_i=1}^{N_m} \vec{\theta}_{m,n_i,select}^*$.

**Step 3: Model selection**

In the literature, model selection is typically performed by the likelihood ratio test combing with Bayesian methods, where competing models are ranked by the ratio of their posterior probabilities (Vyshemirsky & Girolami, 2008). However, since marginal likelihoods cannot be evaluated analytically for CF models, deriving exact posterior distributions is also impossible. Instead, we establish a probabilistic model selection approach based on the distance function in *Eq.* (5).

Firstly, all particles from all models are merged and further selected with their corresponding particle scores:

$$\Theta_{merge} = \cup_{m=1}^{M} \Theta_m(N), \tag{6}$$

where $\Theta_{merge}$ represents the merged particle set. Then we sort the score of particles in $\Theta_{merge}$ and select the best $N^A$ particles as the learned result, denoted by $\Theta_{hybrid}$. Note that we need to set $N^A \gg N$ to guarantee the over-representing issue is not prominent. Since $\Theta_{hybrid}$ may contain the particles from any CF models, we can calculate the percentage of particles selected from certain model to see its impact, denoted by:

$$\beta_m = \frac{1}{N^A} |\Theta_m \cap \Theta_{hybrid}|. \tag{7}$$

It is intuitive that $\beta_m$ is the estimated probability of each model $m$ being selected in the hybrid model based on the estimated posterior distributions.

*2.4. Stochastic and deterministic metrics*

Given the learned hybrid particles $\Theta_{hybrid}$, we evaluate the performance of the learned hybrid stochastic CF model in reproducing CF behaviors. Specifically, we aim to evaluate the distribution-wise goodness-of-fit, in addition to the deterministic assessment based on the particle score.

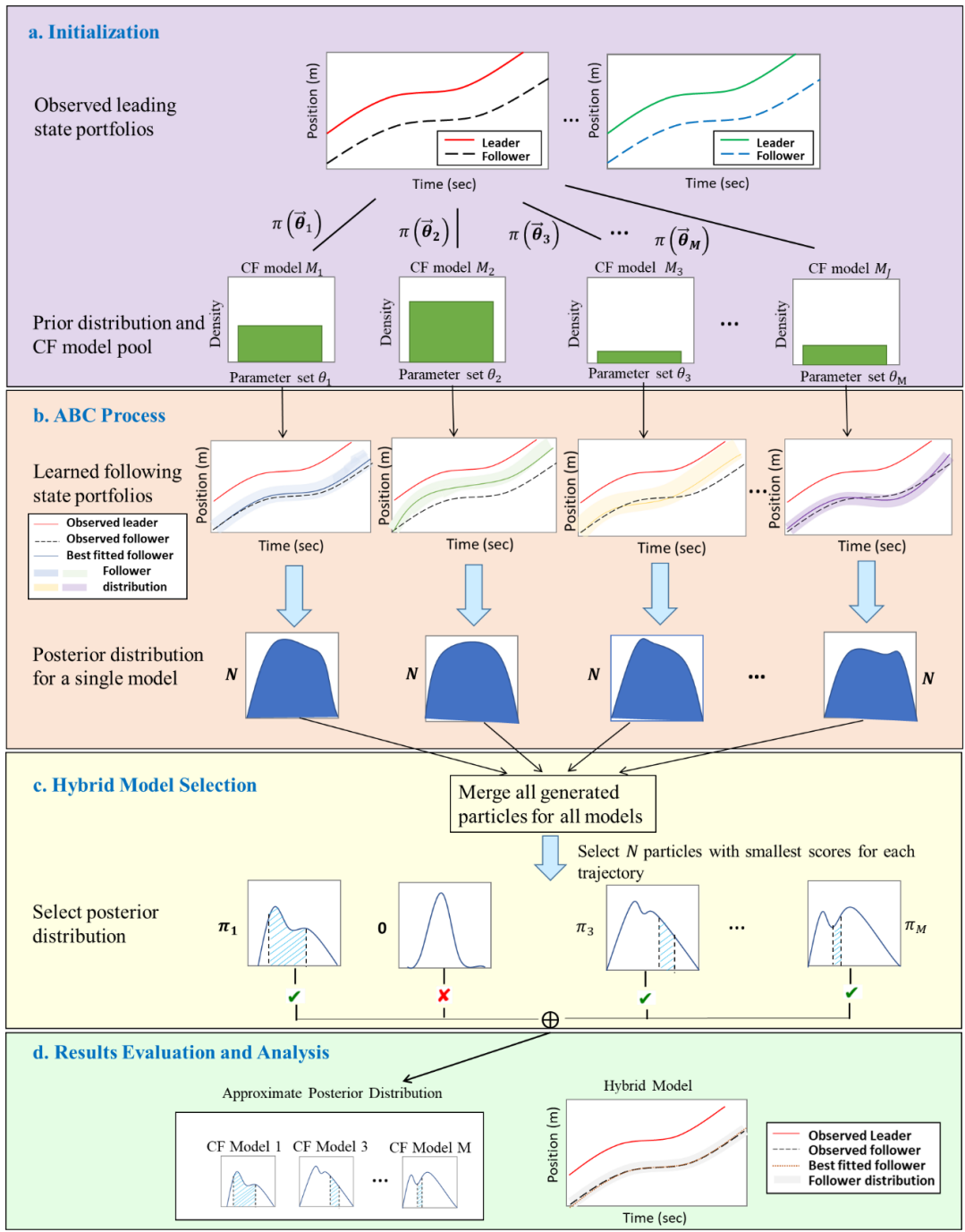

**Fig. 2 General framework of ABC with model selection on CF learning.** Inset shows the four main procedures: **a** Initialization work including defining CF model candidates and corresponding prior distribution. **b** Apply ABC – RS (Beaumont et al., 2002) process for each CF model independently. **c** Conduct a probabilistic hybrid model selection based on predefined scores and acquire estimated posterior distribution. **d** Evaluate the goodness of fit of learned parameter posterior distributions via multiple measures.

To measure the trajectory-level goodness-of-fit in a stochastic fashion, we introduce the Wasserstein distance (WS). The WS distance is widely used to measure the distance of two probabilistic measures, achieved through the solution of a linear programming problem pertaining to optimal transport. There are a few benefits of using WS distance over other distance measures like the Euclidean distance. For example, it can handle distributions that have heavy tails and is more resistant to outliers. Here, we use WS distance as a way of measuring the similarity between the observed posterior distributions and the learned posterior distributions, which can be formulated as:

$$W(\mathbb{y}_f, \mathbb{y}_l, \Theta_{hybrid}) = \inf_{\{\gamma_{i^*,\vec{\theta}^*}\}} \sum_{i^* \in \{1,2,\dots,I\}, \vec{\theta}^* \in \Theta_{hybrid}} \gamma_{i^*,\vec{\theta}^*} \begin{bmatrix} \alpha_1 \|\vec{p}_{f,i^*,lrn}(\vec{\theta}^*, \vec{Y}_{l,i^*,obs}) - \vec{p}_{f,i^*,obs}\| + \\ \alpha_2 \|\vec{v}_{f,i^*,lrn}(\vec{\theta}^*, \vec{Y}_{l,i^*,obs}) - \vec{v}_{f,i^*,obs}\| + \\ \alpha_3 \|\vec{a}_{f,i^*,lrn}(\vec{\theta}^*, \vec{Y}_{l,i^*,obs}) - \vec{a}_{f,i^*,obs}\| \end{bmatrix},$$ (8)

s.t.

$$\sum_{\vec{\theta}^* \in \Theta_{hybrid}} \gamma_{i^*,\vec{\theta}^*} = \frac{1}{I}, \forall i^*,$$ (8a)

$$\sum_{i^*=1}^{I} \gamma_{i^*,\vec{\theta}^*} = \frac{1}{|\Theta_{hybrid}|}, \forall \vec{\theta}^*.$$ (8b)

where $\gamma_{i^*,\vec{\theta}^*} \in [0,1]$ is a joint probability to be determined for each $i^*$ and $\vec{\theta}^*$, whose marginals are constrained by Eq. (8a) (state portfolio constraint) and Eq. (8b) (particle constraint). We explicitly write out the state portfolio component simulated by particle $\vec{\theta}^*$ given a leading vehicle $i^*$, e.g., vehicle position as $\vec{p}_{f,i^*,sim}(\vec{\theta}^*, \vec{Y}_{l,i^*,obs})$.

To be more reliable to extreme values, we further refine the WS distance by partially matching the two distribution and define $\beta$ −Wasserstein (WS) distance as follows:

$$W_\beta(\mathbb{y}_f, \mathbb{y}_l, \Theta_{hybrid}) = \inf_{\{\gamma_{i^*,\vec{\theta}^*}\}} \sum_{i^* \in \{1,2,\dots,I\}, \vec{\theta}^* \in \Theta_{hybrid}} \gamma_{i^*,\vec{\theta}^*} \begin{bmatrix} \alpha_1 \|\vec{p}_{f,i^*,sim}(\vec{\theta}^*, \vec{Y}_{l,i^*,obs}) - \vec{p}_{f,i^*,obs}\| + \\ \alpha_2 \|\vec{v}_{f,i^*,sim}(\vec{\theta}^*, \vec{Y}_{l,i^*,obs}) - \vec{v}_{f,i^*,obs}\| + \\ \alpha_3 \|\vec{a}_{f,i^*,sim}(\vec{\theta}^*, \vec{Y}_{l,i^*,obs}) - \vec{a}_{f,i^*,obs}\| \end{bmatrix},$$ (9)

where $\gamma_{i^*,\vec{\theta}^*} \in [0,1]$ for each $i^*$ and $\vec{\theta}^*$ is a coupling with the following two marginal distribution constraints:

$$\sum_{\vec{\theta}^* \in \Theta_{hybrid}} \gamma_{i^*,\vec{\theta}^*} = \frac{1}{I}, \forall i^*,$$ (9a)

$$\sum_{i^*=1}^{I} \gamma_{i^*,\vec{\theta}^*} \geq \beta \cdot \frac{1}{|\Theta_{hybrid}|}, \forall \vec{\theta}^*.$$ (9b)

*Eq.* (9a) and *Inequality* (9b) are state portfolio constraint and particle constraint, respectively, where $\beta \in (0,1)$ is the percentage of samples selected from each distribution for matching. If $\beta$ is 1, it is the original WS distance, where all samples of both distributions are considered to calculate the distance. The lower value of $\beta$ reflects a higher degree of screening out long tails and outliers.

Further, we define the minimum distance by dropping the particle constraint (i.e., *Inequality* (9b)), i.e., $\beta = 0$. The minimum distance aims to be more inclusive and measure the goodness-of-fit by selecting the smallest score for each state portfolio, permitting the possibility of a single particle being chosen multiple times.

# 3. Composition of Hybrid Model

Since considering all CF models and control algorithms is not feasible or insightful, eight models have been carefully selected, including four well-known HDV CF models and four state-of-the-art AV controllers. For HDVs, contemporary statistical physics-based models, OVM (Bando et al., 1995) and IDM (Kesting et al., 2010), are selected. Additionally, two CF models that extend the OVM have also been included: the Generalized Force Model (GFM) (Helbing & Tilch, 1998) and the Full Velocity Difference Model (FVDM) (Jiang et al., 2001). Notably, GFM addresses the issue of unrealistic high acceleration present in OVM, while FVDM considers both positive and negative velocity differences to describe CF behaviors more comprehensively, especially in cases when the speed of leading vehicle is faster than that of following vehicle. These models are known for theoretical soundness, good agreement with real data, and ability to reproduce key traffic features (Saifuzzaman et al., 2015). Detailed notations and formulas for the selected HDV CF models can be found in Tables 1-3 in Appendix 1. With regard to AVs, CF controllers can mainly differ in three aspects: (1) spacing policy (e.g., constant time gap (CTG) (Swaroop & Hedrick, 1996), constant spacing (CS) (Swaroop & Hedrick, 1996) ); (2) controller type (e.g., linear (Zhou et al., 2020), MPC (Zhou et al., 2019) ); and (3) approximation of vehicle dynamics (e.g., second-order (Zhou et al., 2017) or third-order dynamics (Zhou et al., 2020a) ). After thorough consideration, lower-order linear feedback controller with constant time gap policy (LLCTG) (Swaroop et al., 1994), lower-order linear feedback controller with constant spacing policy (LLCS) (Swaroop et al., 1994), higher-order (HL) linear feedback controller (Zhou et al., 2020), and model predictive controller (MPC) (Zhou et al., 2019) have been selected. More information can be found in Appendix 1.

Notably our ABC-based framework does not require a specific distribution for a CF model parameter, as the posterior distribution is approximated in a numerical fashion. Thus, we assume a simple, uniform prior distribution within a reasonable range for each parameter reported in the literature (Bando et al., 1995; Helbing & Tilch, 1998; Jiang et al., 2001; Kesting et al., 2010; Swaroop et al., 1994; Zhou et al., 2019, 2020). The lower bounds and upper bounds of learning parameter sets for all models are included in Table 1.

Table 1 Parameters and corresponding prior distribution bounds for each model

| Parameter | Lower bound | Upper bound |
|---|---|---|
| **OVM** | | |
| Sensitive parameter, $\kappa\ (sec^{-1})$ | 0.5 | 2 |
| Speed factor, $v_1\ (m/s)$ | 5 | 8 |
| Speed factor, $v_2\ (m/s)$ | 20 | 25 |
| Form factor, $c_1 (m^{-1})$ | 0.05 | 0.2 |
| Form factor, $c_2 (m^{-1})$ | 1.5 | 1.7 |
| **GFM** | | |
| Proportionality factor, $K$ | 0 | 2 |
| Sensitivity factor, $\lambda$ | 0 | 2 |
| Speed factor, $v_1\ (m/s)$ | 0 | 10 |
| Speed factor, $v_2\ (m/s)$ | 0 | 30 |
| Form factor, $c_1 (m^{-1})$ | 0 | 0.2 |
| Form factor, $c_2 (m^{-1})$ | 1 | 2 |
| **FVDM** | | |
| Relaxation time, $\tau\ (s^{-1})$ | 600 | 2000 |
| Sensitivity parameter, $\lambda\ (s)$ | 0 | 2 |
| Speed $V_1 (m/s)$ | 0 | 40 |
| Speed $V_2 (m/s)$ | 0 | 40 |
| Interaction length, $l_{int}\ (m)$ | 0 | 40 |
| Unitless parameter, $\beta$ | 0 | 40 |
| **IDM** | | |
| Desired speed, $v_{max}\ (m/s)$ | 20 | 40 |
| Desired time gap, $T\ (s)$ | 0.8 | 2.5 |
| Minimum gap (jam distance), $s_0\ (m)$ | 0.5 | 3 |
| Maximum acceleration, $a\ (m/s^2)$ | 0.5 | 2 |
| Desired deceleration, $b\ (m/s^2)$ | 1 | 4 |
| Free acceleration exponent, $\delta$ | 2 | 5 |
| **LLCTG** | | |
| Desired time gap, $\tau^*(s)$ | 0.8 | 1.2 |
| Spacing deviation feedback gain, $k_s$ | 0.3 | 2.3 |
| Speed difference feedback gain, $k_v$ | 0.3 | 2.3 |
| Standstill distance, $l(m)$ | 1 | 11 |
| **LLCS** | | |
| Desired spacing, $s_0(m)$ | 5 | 25 |
| Spacing deviation feedback gain, $k_s$ | 0.3 | 2.3 |
| Speed difference feedback gain, $k_v$ | 0.3 | 2.3 |
| **HL** | | |
| Desired time gap, $\tau^*\ (s)$ | 0.8 | 1.2 |
| Actuation lag, $TT\ (s)$ | 0.1 | 0.5 |
| Spacing deviation feedback gain, $k_s$ | 0.1 | 2.3 |
| Speed difference feedback gain, $k_v$ | 0.1 | 2.3 |
| Acceleration feedback gain, $k_a$ | -3 | 0 |
| Standstill distance, $l(m)$ | 3 | 8 |
| **MPC** | | |
| Desired time gap, $\tau^*\ (s)$ | 0.6 | 1.4 |
| Comfort and fuel consumption, $R$ | 0.3 | 1.7 |
| Control efficiency coefficient, $\alpha$ | 0.3 | 1.7 |
| Standstill distance, $l(m)$ | 3 | 7 |
| Deceleration limit, $a_{min}\ (m/s^2)$ | -5 | -3 |
| Acceleration limit, $a_{max}(m/s^2)$ | 3 | 5 |

# 4. Experiments and Learning Results

*4.1. Data sources*

To train the proposed stochastic hybrid model, two datasets have been selected to train the model for HDVs and AVs. Specifically, the widely used NGSIM dataset has been selected to train the model for HDVs, while the Massachusetts (MA) Experiment dataset (Li et al., 2022) has been selected to train for AVs. Note that the MA experimental dataset has been further categorized into two datasets: CAR MODEL I and CAR MODEL II, representing two different AV controllers. The actual car models are omitted here to avoid potential conflicts of interest. In each dataset, the movements of leading-following vehicle pairs are recorded by sensors that measure the vehicle position, speed, and acceleration.

In particular, we focus on trajectory pairs between 4:00 – 4:15 AM on I-80 for NGSIM. After simple data processing, 150 CF-pairs are randomly selected as our input, each with a 35-second duration. The MA dataset consists of 96 and 64 trajectories for CAR MODEL 1 and CAR MODEL II, respectively. The duration of each trajectory pair is 54.1 seconds for CAR MODEL I and 57.6 seconds for CAR MODEL II. The original field data may also include longitudinal speed and acceleration data. However, due to the limitation of the experiments, only the position data are reliable. Therefore, a finite difference method is applied to numerically calculate the speed and acceleration.

*4.2. Learning results for HDVs*

First, we apply our ABC-based framework to learn HDV CF behaviors using the NGSIM dataset. To mitigate the potential bias inherent in single train-test splits, we employ cross-validation. Specifically, the dataset is evenly divided into three parts. We train the model based on two of these parts and evaluate its performance on the remaining part, thereby establishing a training-to-testing ratio of the 2:1. This process is iterated three times, and the results are aggregated by calculating the average of chosen metrics. During the training, 1 million particles (i.e., parameter sets) were sampled from the assumed prior distributions for each model and accepted/rejected based on the predefined distance function.

The training result for HDVs is shown in Fig. 3a. It can be observed that the HDV CF models (GFM, FVDM, OVM and IDM) overshadow the AV controllers (MPC and LLCTG) in the hybrid model, making up more than 97% of accepted particles. Among them, GFM has the highest share of approximately 65%. GFM, FVDM, and OVM, which belong to the same model family, all take nonnegligible shares, indicating that this model family can effectively describe the HDV CF behavior. In contrast, HL and LLCS are completely dropped in the hybrid model, suggesting that the CF behavior of AVs is different from the behavior of human drivers.

For a more direct comparison of the selected models, we further examined the pair-wise relative likelihood of one CF model fitting the observed behavior better than the other. Specifically, we replicated the training process for each pair of CF models (out of 28 enumerated pairs). Since only two models are compared, the model with more than 50% is considered preferable. Fig. 3d presents the pair-wise comparison in the form of a heat map, where the value indicates the relative likelihood (i.e., proportion of accepted particles) for one model (row) against the other (column). The result further confirms that GFM, FVDM, and OVM are strongly favored, with GFM showing the clearest preference when compared to the other models one-on-one.

For the goodness of fit evaluation at the vehicle trajectory level, we compare the deviation between the learned (based on accepted particles) and observed vehicle positions. To highlight the performance of the proposed hybrid model, we compare with the best single CF model for each dataset: GFM (for NGSIM), HL (for CAR MODEL I), and IDM (for CAR MODEL II) (refer to Fig. 3). Fig. 4a shows an example of the evolution of position error for the hybrid model, as compared to these best single models (Fig. 4d). In these figures, we plot the error evolution for the 5% best fitted particles (red/blue/green/purple) and all selected particles (light green). Comparing these figures, we observe that the hybrid model has lower errors in general than a single model and shows a stable trend over time.

Further, we compare its training results against those of individual models that are stochastically learned with the incorporation of only one CF model in the learning process. The evaluation metrics consist of two types: (1) absolute errors and (2) distribution-wise similarity. Specifically, (1) comprises errors in average position, average speed, and average acceleration. More importantly, (2) is measured using the goodness-of-fit metrics, Wasserstein (WS) Distance, 0.15-WS Distance ($\beta = 0.15$), and minimum distance, specially designed in this study. In principle, these distances measure the deviations between the state portfolios generated based on accepted particles and the corresponding observed ones, using different constraints and weights on the position, speed, and acceleration. Detailed numerical results of these three cross-validation trials are included in Appendix Tables 4-6 in Appendix 2. Here, we focus on the general performance trend for each model across the six metrics. To address the scale inconsistency across the metrics, a linear normalization step is taken. Fig. 5 visually presents the results for the NGSIM dataset through a series of hexagonal-based diagrams. Within each hexagon, the aforementioned six metrics are positioned as six vertices, constrained within the normalized range of 0 to 1. A larger shaded area signifies a higher level of performance. The results reveal that among the single models, the GFM model performs the best in general, but not in all metrics, and clear deficiency is notable. The hybrid model exhibits the largest shaded (blue) area when compared to other single models, showcasing superior performance across all metrics. This highlights the hybrid model can better capture HDV CF behavior considering stochasticity than other conventional CF models.

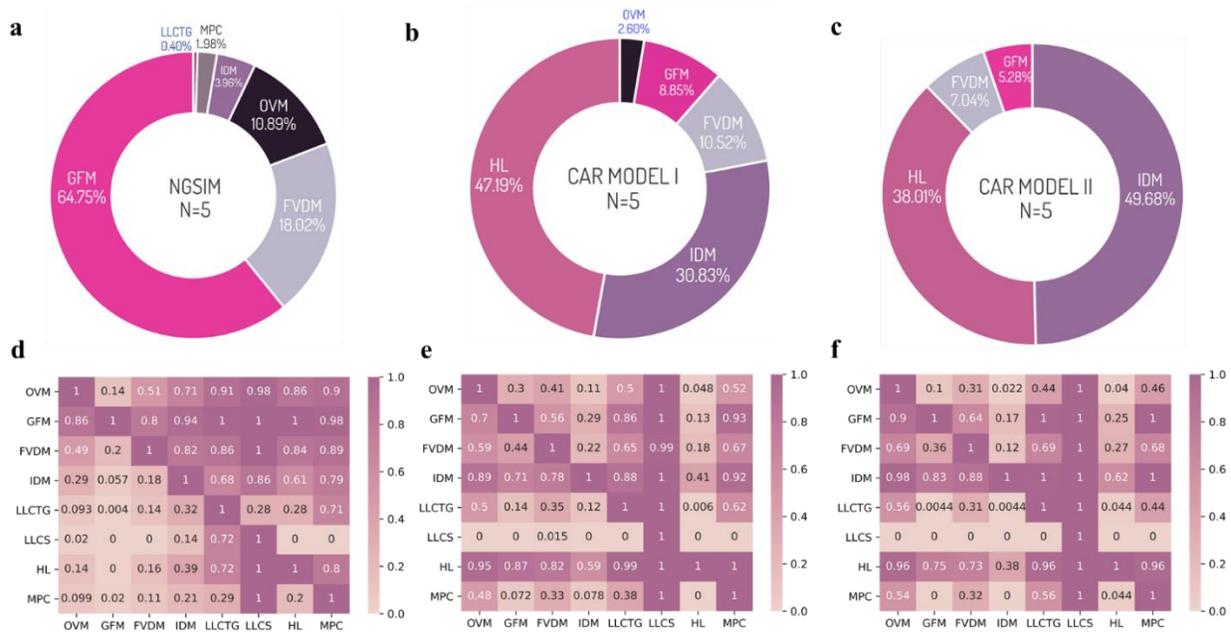

**Fig. 3 Training results. a** Hybrid model distribution – NGSIM. **b** Hybrid model distribution - CAR MODEL I. **c** Hybrid model distribution – CAR MODEL II. **d** Heatmap of pairwise model selection probabilities – NGSIM. **e** Heatmap of pairwise model selection probabilities – CAR MODEL I. **f** Heatmap of pairwise model selection probabilities – CAR MODEL II. The selected number of particles in testing set is 5. (e.g., row 2, column 1 for **d**, GFM and OVM are the two models selected for training, and 86% of particles are selected from GFM, while the remaining 14% are from OVM. Hence, this tells that GFM significantly dominates OVM in the hybrid model.)

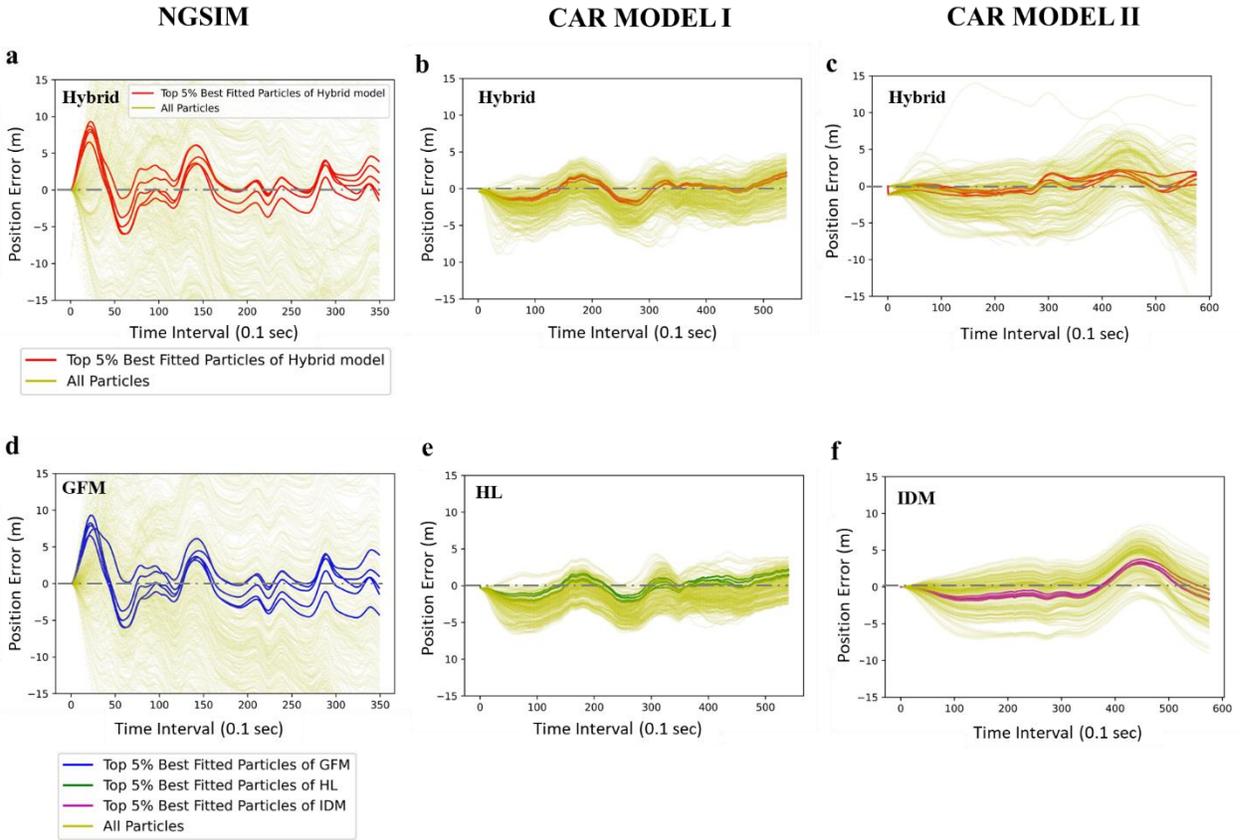

**Fig. 4 Goodness of fit evaluation results. a** Hybrid model position error evolution – NGSIM (#27). **b** Hybrid model position error evolution – CAR MODEL I (#1). **c** Hybrid model position error evolution – CAR MODEL II (#8). **d** Single model position error evolution (GFM) – NGSIM (#27). **e** Single model position error evolution (HL, IDM) – CAR MODEL I (#1). **f** Single model position error evolution (HL, IDM) – CAR MODEL II (#8). Note: the numbers in parentheses indicate a specific sampled state portfolio in the testing set.

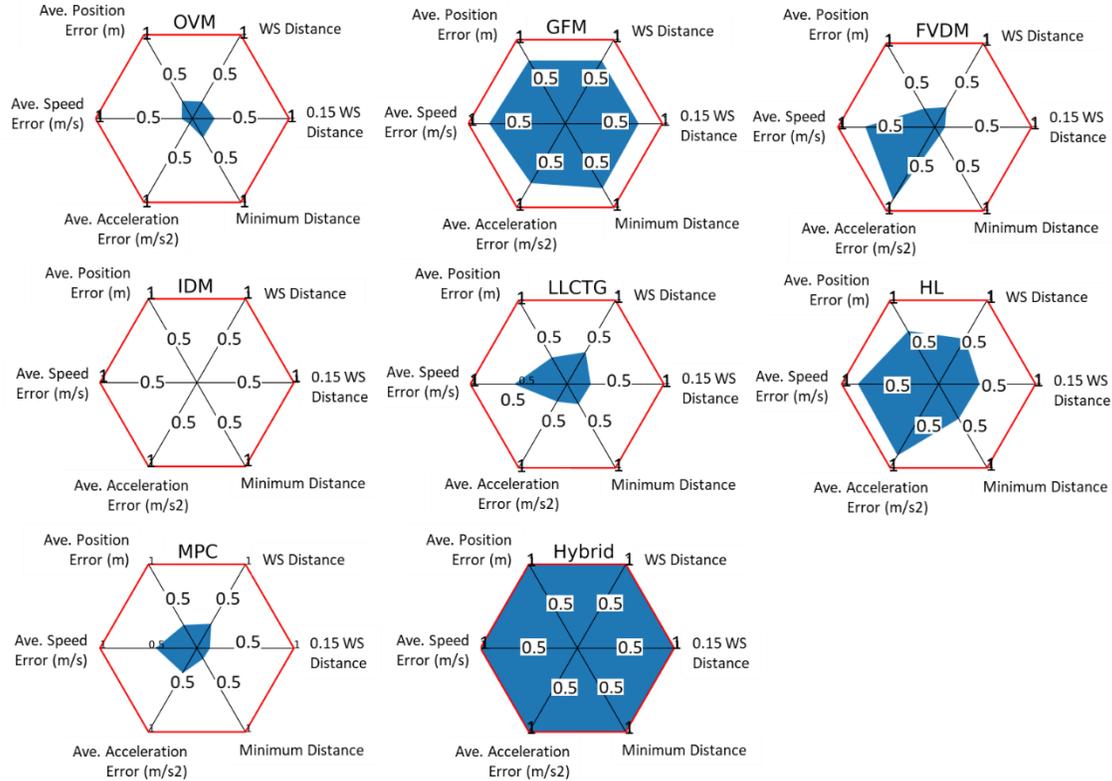

**Fig. 5 Hexagonal based Multiple Metric Performance Evaluation for HDV.**

## 4.3. Learning results for AVs

Here we turn our attention to learning AV CF behavior from the CAR MODEL I dataset. The results in Fig. 3b show that two AV controllers, LLCS and MPC, are not selected in the hybrid model. Of the remaining models, HL has the largest share of the distribution at 47.19%, and IDM (a HDV CF model) also takes up a sizable share at 30.83%. The lack of dominance by HL suggests that the CF control algorithm of the AV in this dataset is similar to, but not necessarily the same as, HL. Given the limited controller information available, there exists the possibility of missing the actual CF controller for CAR MODEL I. Therefore, selecting HL (deterministically) to approximate the behavior of the AV can give us erroneous insights. The hybrid model, however, fills this gap by identifying a set of models that can together approximate the AV behavior in the absence of controller knowledge. The heatmap in Fig. 3e further corroborates the findings in Fig. 3b that the two models (HL and IDM) show strong preference over the other models, but HL shows mild preference over IDM. Figs. 4b & 4e depict an example of the deviation evolution between the observed and learned position. The position generated by the hybrid model's top 5% of the best fitted particles exhibits better accuracy, as the position errors display a more centralized trend to zero compared to the best single model (HL).

The learning results from the CAR MODEL II dataset, shown in Fig. 3c, also show split preference with no clear dominance, particularly between IDM and HL, with IDM accounting for the largest proportion. This suggests a great possibility that the true controller of CAR MODEL II might not be included in the CF candidate pool. When comparing the training results with CAR MODEL I, the composition of the hybrid model is similar, except for the exclusion of OVM. The corresponding heatmap in Fig. 1f also displays high similarity with Fig. 3e. The evaluation results in Figs. 4c and 4f also demonstrate comparable error ranges with the CAR MODEL I dataset. However, we observe that the CF behavior may not be well approximated by a single IDM due to its larger error than the hybrid model. Therefore, when true controller is absent, the hybrid model obtained by concatenation can generate a more accurate CF behavior description.

Fig. 6 illustrates the general performances across the six metrics for each model and the hybrid model for two AV datasets. The results are weighted by the sample size (i.e., number of state portfolios) in each dataset. The hybrid model demonstrates the best overall performance, though some exceptions in certain metrics are observed (e.g., average acceleration). Among the single models, HL shows the best overall performance. The results demonstrate the effectiveness of the hybrid model for capturing the behavior of AVs, particularly when the controller information is unavailable.

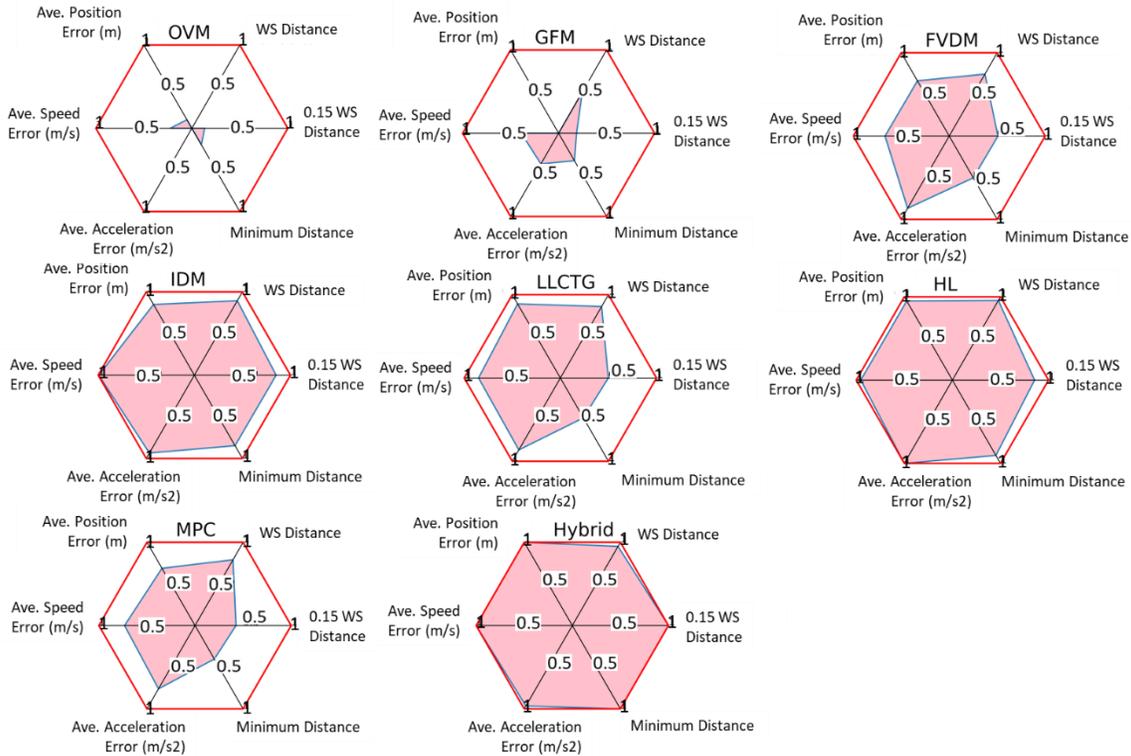

**Fig. 6 Hexagonal based Multiple Metric Performance Evaluation for AV.**

## 5. Conclusion and Discussion

Learning the real-world CF behavior has been challenging due to inherent stochasticity and nonlinearity that arise from driver heterogeneity. Traditional approaches that rely on a single (deterministic) model often fail to capture these characteristics sufficiently, leaving room for persisting debates about the best CF model. The proposed hybrid model based on stochastic learning of ABC addresses these challenges by integrating different CF models in a systematic and stochastic fashion. The evaluation of the proposed framework using two different datasets demonstrated the ability to learn the CF behavior while accounting for stochasticity, outperforming traditional CF models. In addition, when the actual CF model or controller is absent from the candidate pool, the hybrid model is still able to replicate the CF behavior by probabilistically concatenating several models.

Our learning framework is highly adaptable to various datasets and the CF model pool for better generalizability. The framework is also flexible in assuming prior distributions of model parameters (ranges and shapes), through it can affect the convergence. However, it should be noted that the learning performance depends on the data quality as with any data-driven approaches. Further, if the CF model pool does not include the true CF (control) model or similar models, the hybrid model would lack interpretability as well as ability to replicate the observed behavior. Future research is needed to design a good pool of candidate models in the event that the true model is unknown. For example, preliminary learning may be conducted to identify the most promising model family.

The proposed framework has several important implications for traffic modeling. Its ability to systematically draw from multiple models can be particularly useful in traffic simulations, where generating traffic flow mimicking the real-world traffic is of high priority. Further, its hybrid approach of integrating multiple physics-based CF models via data-driven ABC method, brings notable strengths. It is able to handle intra- and inter-driver heterogeneities while preserving physical interpretability. Compared to pure data-driven methods, our approach is less data-intensive and better equipped to handle safety-critical corner cases(Feng et al., 2023), especially when data availability is limited. More broadly, the proposed stochastic learning of CF behavior is just one potential application. Our approach can be applied to various learning problems in traffic contexts, such as learning of lane changing and merging behavior, and AV behavior under diverse traffic scenarios(Feng et al., 2021), and beyond where high degrees of non-linearity and stochasticity in behavior are expected.

## Contributions

J.J., Y.Z., X.W., and S.A., conceptualized and designed the study. J.J. and X.W. processed the data and wrote codes of ABC – hybrid model framework. J.J prepared for the draft manuscript. X.W., Y.Z., and S.A. provided feedback during the manuscript revisions and results discussion. All authors approved the final version of the manuscript. S.A. approved the submission and accepted responsibility for the overall integrity of the paper.

## Acknowledgements

This research is supported by the National Science Foundation CNS #1739869 and CMMI #2129765. We also sincerely thank Dr. Danjue Chen for sharing the Massachusetts AV Experiment data.## Appendix

**Appendix 1: Notations and Formulas for CF Models and Controllers**

Table A1 shows the corresponding variable and parameter notations of CF models and controllers in the candidate pool as described in Section 3 of the main manuscript.

Table A2 and Table A3 present the detailed formulations of CF models and controllers, respectively. They employ the variables and parameters defined in Table A1. Specifically, for traditional HDV CF models, acceleration is formulated combining with a predefined desired speed or spacing policy. For AV controllers, the system updates itself using the state-space formulation. Therefore, aside from the desired spacing policy, the system state is also defined.

## Appendix Table 1 Variable and parameter notations of CF models and controllers

| HDV Model Notations | Description |
|---|---|
| $p_i(t)$ | Position of vehicle $i$ at time $t$ |
| $L_{i-1}$ | Length of vehicle $i-1$ |
| $v_i(t)$ | Speed of vehicle $i$ at time $t$ |
| $s_i(t)$ | Gap between vehicle $i$ and $i-1$ at time $t$ |
| $v_i^*(s_i(t))$ | Optimal velocity of vehicle $i$ in OVM and GFM |
| $\Delta v_i(t)$ | Speed difference, can be $v_{i-1}(t) - v_i(t)$ or $v_i(t) - v_{i-1}(t)$ depended on model |
| $V(s_i(t))$ | Optimal velocity of vehicle $i$ in FVDM |
| $s^*(t)$ | Desired spacing in IDM |
| **CAV Controller Notations** | **Description** |
| $x_i(t)$ | System state of vehicle $i$ at time $t$ |
| $u_i(t)$ | System input of vehicle $i$ at time $t$, can be viewed as acceleration |
| $x_{i,t}$ | Discretized system state of vehicle $i$ at time $t$ |
| $u_{i,t}$ | Discretized system input of vehicle $i$ at time $t$ |
| $s_i^*(t)$ | Desired spacing |
| $\tau_i^*$ | Desired time headway |
| $l_i$ | Minimum standstill spacing |
| $s_i(t)$ | Actual spacing |
| $\Delta s_i(t)$ | Deviation from the desired spacing |
| $\Delta v_i(t)$ | Relative speed |
| $k_{di}$ | Discretized feedback gains |
| $k_{si}, k_{vi}, k_{ai}$ | Feedback gains for deviation from the desired spacing, relative speed, and acceleration |
| $A_{di}, B_{di}, D_{di}$ | Discretized system weight matrices |
| $a_{i-1}(t)$ | Acceleration of vehicle $i-1$ (leading vehicle) |
| $t_s$ | Control frequency (interval) |
| $s_0$ | A fixed positive value for desired spacing |
| $TT_i$ | Actuation lag for vehicle $i$ to realize the acceleration |
| $J$ | Optimal objective function |
| $Q_i$ | Control efficiency function |
| $R_i$ | Comfort and fuel consumption function |
| $a_{i,min}$ | Lower bound of acceleration |
| $a_{i,max}$ | Upper bound of acceleration |

## Appendix Table 2 Formulations of CF models and controllers

| HDV Model | Desired Speed/Spacing | Acceleration Formulation |
|---|---|---|
| OVM | $v_i^*(s_i(t)) = v_1 + v_2[\tanh(c_1 * (s_i(t)) - c_2]$ <br> where $s_i(t) = p_{i-1}(t) - p_i(t) - L_{i-1}$ | $\frac{dv_i}{dt}(t) = \kappa[v_i^*(s_i(t)) - v_i(t)]$ |
| GFM | Same as OVM | $\frac{dv_i}{dt}(t) = K[v_i^*(s_i(t)) - v_i(t)] + \lambda\Theta(-\Delta v)\Delta v$ <br> where $\Theta := \begin{cases} 1, & -\Delta v > 0 \\ 0, & -\Delta v \leq 0 \end{cases}$ <br> $\Delta v = v_{i-1}(t) - v_i(t)$ |
| FVDM | $V(s_i(t)) = V_1 + V_2 \tanh\left[\frac{s_i(t) - L_{i-1}}{l_{int}} - \beta\right]$ <br> $s_i(t) = p_{i-1}(t) - p_i(t)$ | $\frac{dv_i}{dt}(t) = \frac{1}{\tau}[V(s_i(t)) - v_i(t)] + \lambda\Delta v$ |
| IDM | $s^*(t) = s_0 + v(t) * T + \frac{v(t)\Delta v(t)}{2\sqrt{ab}}$ <br> where $\Delta v(t) = v_i(t) - v_{i-1}(t)$ <br> $s_i(t) = p_{i-1}(t) - p_i(t)$ | $\frac{dv_i}{dt}(t) = a\left[1 - \left(\frac{v_i(t)}{v_{max}}\right)^\delta - \left(\frac{s^*(t)}{s(t)}\right)^2\right]$ |

**Appendix Table 3 Formulations of CF controllers**

| AV Controller | Desired Spacing Policy | System State | State-Space Formulation |
|---|---|---|---|
| LLCTG | $s_i^*(t) = v_i(t) \times \tau_i^* + l_i$ | $x_i(t) = [\Delta s_i(t), \Delta v_i(t)]^T$ where $\Delta s_i(t) = s_i(t) - s_i^*(t)$, $\Delta v_i(t) = v_{i-1}(t) - v_i(t)$ $k_{di} = [k_{si}, k_{vi}]^T$ | $x_{i,t+1} = A_{di} x_{i,t} + B_{di} u_{i,t} + D_{di} a_{i-1,t}$ where $A_{di} = \begin{pmatrix} 1 & t_s \\ 0 & 1 \end{pmatrix}$, $B_{di} = \begin{pmatrix} -t_s \tau_i^* - t_s^2/2 \\ -t_s \end{pmatrix}$, $D_{di} = \begin{pmatrix} t_s + t_s^2/2 \\ t_s \end{pmatrix}$ $u_{i,t} = k_{di} x_{i,t}$ |
| LLCS | $s_i^*(t) = s_0$ | Same as LL | $x_{i,t+1} = A_{di} x_{i,t} + B_{di} u_{i,t} + D_{di} a_{i-1,t}$ where $A_{di} = \begin{pmatrix} 1 & t_s \\ 0 & 1 \end{pmatrix}$, $B_{di} = \begin{pmatrix} -\frac{t_s^2}{2} \\ -t_s \end{pmatrix}$, $D_{di} = \begin{pmatrix} t_s^2/2 \\ t_s \end{pmatrix}$ $u_{i,t} = k_{di} x_{i,t}$ |
| HL | $s_i^*(t) = v_i(t) \times \tau_i^* + l_i$ | $x_i(t) = [\Delta s_i(t), \Delta v_i(t), a_i(t)]^T$, $\dot{a}_i(t) = -\frac{1}{TT_i} a_i(t) + \frac{1}{TT_i} u_i(t)$ $k_{di} = [k_{si}, k_{vi}, k_{ai}]^T$ | $x_{i,t+1} = A_{di} x_{i,t} + B_{di} u_{i,t} + D_{di} a_{i-1,t}$ where $A_{di} = \begin{pmatrix} 1 & t_s & TT_i(\tau_i^* - TT_i)\left(e^{-\frac{t_s}{TT_i}} - 1\right) - t_s \cdot TT_i \\ 0 & 1 & TT_i\left(e^{-\frac{t_s}{TT_i}} - 1\right) \\ 0 & 0 & e^{-\frac{t_s}{TT_i}} \end{pmatrix}$, $B_{di} = \begin{pmatrix} -TT_i(\tau_i^* - TT_i)\left(e^{-\frac{t_s}{TT_i}} + \frac{t_s}{TT_i} - 1\right) - \frac{t_s^2}{2} \\ TT_i\left(1 - e^{-\frac{t_s}{TT_i}}\right) - t_s \\ 1 - e^{-\frac{t_s}{TT_i}} \end{pmatrix}$, $D_{di} = \begin{pmatrix} \frac{t_s^2}{2} \\ t_s \\ 0 \end{pmatrix}$ |
| MPC | $s_i^*(t) = v_i(t) \times \tau_i^* + l_i$ | $x_i(t)$ same as HL $\min J = (x_{i,t})^T \cdot Q_i \cdot x_{i,t} + R_i \cdot (u_{i,t-1})^2$ s.t. $a_{i,min} \leq u_i(t) \leq a_{i,max}$ where $Q_i = \begin{bmatrix} 1 & 0 \\ 0 & \alpha \end{bmatrix}, R_i > 0$ | $x_{i,t+1} = A_{di} x_{i,t} + B_{di} u_{i,t} + D_{di} a_{i-1,t}$ $A_{di} = \begin{pmatrix} 1 & t_s \\ 0 & 1 \end{pmatrix}$, $B_{di} = \begin{pmatrix} -\tau^* \cdot t_s - t_s - \frac{t_s^2}{2} \\ -t_s \end{pmatrix}$, $D_{di} = \begin{pmatrix} t_s + \frac{t_s^2}{2} \\ t_s \end{pmatrix}$ |

## Appendix 2: Supplementary learning results

1. *Sensitivity Analysis of the Number of Selected Particles*

Appendix Fig. 1 illustrates the learned hybrid model distributions with varying numbers of selected particles, as obtained from the NGSIM dataset. In comparison with Fig. 3a in the main manuscript, the learning results are relatively robust against the number of particles being selected.

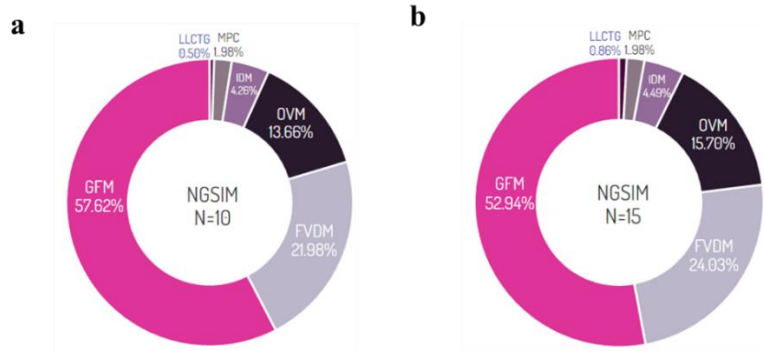

**Appendix Fig. 1. Hybrid model distribution – NGSIM. a** The selected number of particles (N) = 10. **b** The selected number of particles (N) = 15.

Similarly, the hybrid model distribution trained by CAR MODEL I is shown in Appendix Fig. 2. Compared with Fig. 3b in the main manuscript, the overall composition of the hybrid model remains consistent. However, with an increase in the number of particles being selected for each state portfolio, a quite small proportion of LLCTG is also included.

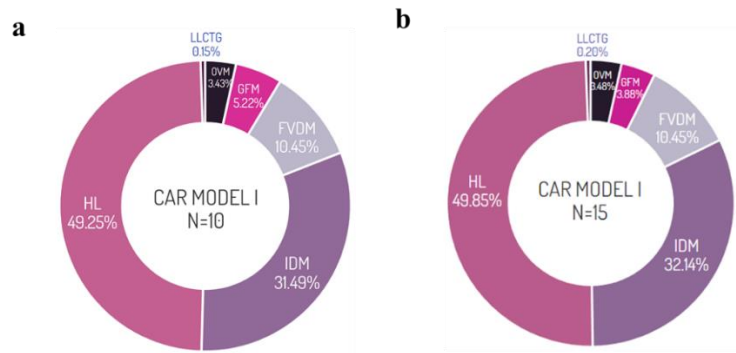

**Appendix Fig. 2. Hybrid model distribution – CAR MODEL I. a** The selected number of particles (N) = 10. **b** The selected number of particles (N) = 15.

Appendix Fig. 3 shows the hybrid model distribution for CAR MODEL II. Same conclusion can be drawn that hybrid model distribution is relatively stable with the changing of selected particle numbers for each state portfolio.

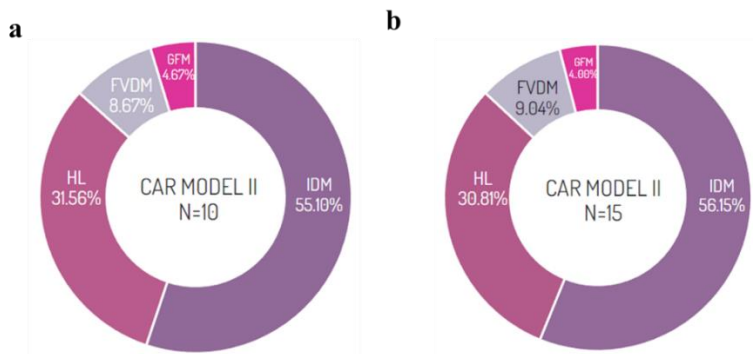

**Appendix Fig. 3. Hybrid model distribution – CAR MODEL II. a** The selected number of particles (N) = 10. **b** The selected number of particles (N) = 15.

2. *Errors and stochastic solution distances*

We bold the minimum value in each column to indicate the model with the optimal performance for each metric.

**Appendix Table 4 Errors and stochastic solution distances: single model vs. hybrid model for NGSIM dataset**

| Model | Average position error ($m$) | Average speed error ($m/s$) | Average acceleration error ($m/s^2$) | Minimum distance | $0.15\beta$-WS distance | WS distance |
|---|---|---|---|---|---|---|
| OVM | 25.532 | 2.923 | 1.520 | 12.041 | 12.074 | 14.168 |
| GFM | 14.703 | 2.050 | 1.287 | 6.498 | 6.789 | 8.477 |
| FVDM | 25.308 | 2.137 | 1.232 | 13.309 | 13.335 | 13.831 |
| IDM | 30.553 | 3.051 | 1.470 | 14.261 | 14.328 | 16.215 |
| LLCTG | 23.386 | 2.367 | 1.468 | 11.908 | 11.934 | 12.330 |
| LLCS | 35.649 | 2.857 | 1.501 | 19.197 | 19.209 | 19.299 |
| HL | 17.100 | 1.989 | 1.237 | 10.113 | 10.153 | 10.664 |
| MPC | 24.218 | 2.520 | 1.437 | 13.030 | 13.050 | 13.266 |
| **Hybrid** | **9.575** | **1.763** | **1.180** | **4.088** | **4.219** | **5.809** |

**Appendix Table 5 Errors and stochastic solution distances: single model vs. hybrid model for CAR MODEL I**

| Model | Average position error ($m$) | Average speed error ($m/s$) | Average acceleration error ($m/s^2$) | Minimum distance | $0.15\beta$-WS distance | WS distance |
|---|---|---|---|---|---|---|
| OVM | 3.287 | 0.637 | 1.838 | 1.662 | 1.678 | 2.217 |
| GFM | 3.293 | 0.550 | 1.813 | 1.788 | 1.791 | 1.807 |
| FVDM | 2.194 | 0.564 | 1.794 | 1.465 | 1.476 | 1.638 |
| IDM | 2.001 | 0.429 | 1.789 | 1.352 | 1.360 | 1.524 |
| LLCTG | 1.897 | 0.511 | 1.790 | 1.383 | 1.389 | 1.463 |
| LLCS | 3.644 | 0.686 | 1.817 | 2.413 | 2.419 | 2.531 |
| HL | 1.940 | 0.511 | **1.783** | 1.316 | 1.332 | 1.493 |
| MPC | 2.004 | 0.537 | 1.797 | 1.440 | 1.441 | 1.530 |
| **Hybrid** | **1.839** | **0.503** | 1.785 | **1.269** | **1.286** | **1.436** |

**Appendix Table 6 Errors and stochastic solution distances: single model vs. hybrid model for CAR MODEL II**

| Model | Average position error ($m$) | Average speed error ($m/s$) | Average acceleration error ($m/s^2$) | $0.15\beta$-WS distance | Minimum distance | WS distance |
|---|---|---|---|---|---|---|
| OVM | 4.145 | 0.716 | 0.903 | 1.940 | 2.098 | 3.790 |
| GFM | 4.764 | 1.250 | 0.879 | 1.251 | 1.656 | 2.736 |
| FVDM | 3.475 | 0.443 | 0.805 | 1.750 | 1.783 | 2.041 |
| IDM | 2.826 | 0.310 | 0.802 | 1.218 | 1.299 | 1.676 |
| LLCTG | 2.834 | 0.572 | 0.817 | 2.040 | 2.048 | 2.163 |
| LLCS | 11.154 | 0.772 | 0.856 | 6.098 | 6.107 | 6.163 |
| HL | 2.357 | 0.332 | **0.802** | 1.180 | 1.367 | **1.453** |
| MPC | 3.804 | 0.577 | 0.824 | 2.066 | 2.073 | 2.258 |
| **Hybrid** | **2.308** | **0.305** | 0.805 | **1.079** | **1.154** | 1.738 |

Swaroop, D., & Hedrick, J. K. (1996). String stability of interconnected systems. *IEEE Transactions on Automatic Control*, *41*(3), 349–357. https://doi.org/10.1109/9.486636

SWAROOP, D., HEDRICK, J. K., CHIEN, C. C., & IOANNOU, P. (1994). A Comparison of Spacing and Headway Control Laws for Automatically Controlled Vehicles. *Vehicle System Dynamics*, *23*(1), 597–625. https://doi.org/10.1080/00423119408969077

Tavaré, S., Balding, D. J., Griffiths, R. C., & Donnelly, P. (1997). Inferring coalescence times from DNA sequence data. *Genetics*, *145*(2), 505–518.

Toni, T., Welch, D., Strelkowa, N., Ipsen, A., & Stumpf, M. P. H. (2009). Approximate Bayesian computation scheme for parameter inference and model selection in dynamical systems. *Journal of the Royal Society Interface*, *6*(31), 187–202. https://doi.org/10.1098/rsif.2008.0172

Treiber, M., Kesting, A., & Helbing, D. (2010). Three-phase traffic theory and two-phase models with a fundamental diagram in the light of empirical stylized facts. *Transportation Research Part B: Methodological*, *44*(8), 983–1000. https://doi.org/https://doi.org/10.1016/j.trb.2010.03.004

Vyshemirsky, V., & Girolami, M. A. (2008). Bayesian ranking of biochemical system models. *Bioinformatics*, *24*(6), 833–839. https://doi.org/10.1093/bioinformatics/btm607

Wiedemann, R. (1974). Simulation des Strassenverkehrsflusses. *Transportation Research Board.*

Zhou, Y., Ahn, S., Chitturi, M., & Noyce, D. A. (2017). Rolling horizon stochastic optimal control strategy for ACC and CACC under uncertainty. *Transportation Research Part C: Emerging Technologies*, *83*, 61–76. https://doi.org/https://doi.org/10.1016/j.trc.2017.07.011

Zhou, Y., Ahn, S., Wang, M., & Hoogendoorn, S. (2020). Stabilizing mixed vehicular platoons with connected automated vehicles: An H-infinity approach. *Transportation Research Part B: Methodological*, *132*, 152–170. https://doi.org/10.1016/j.trb.2019.06.005

Zhou, Y., Jafarsalehi, G., Jiang, J., Wang, X., Ahn, S., & Lee, J. D. (2022). *Stochastic Calibration of Automated Vehicle Car-Following Control: An Approximate Bayesian Computation Approach*. https://doi.org/http://dx.doi.org/10.2139/ssrn.4084970

Zhou, Y., Wang, M., & Ahn, S. (2019). Distributed model predictive control approach for cooperative car-following with guaranteed local and string stability. *Transportation Research Part B: Methodological*, *128*, 69–86. https://doi.org/10.1016/j.trb.2019.07.001
25